\documentclass[letterpaper, 10pt, conference]{ieeeconf}
\IEEEoverridecommandlockouts                             
\usepackage{graphics} 
\usepackage{epsfig} 
\usepackage{times} 
\usepackage{amsmath,mathtools} 
\usepackage{yhmath}
\usepackage{amssymb}  
\usepackage{centernot}
\usepackage{pbox}
\usepackage{svg}
\usepackage{bm}

\usepackage{url}
\usepackage{hyperref}

\usepackage{xcolor}

\usepackage{color, colortbl}
\definecolor{Gray}{gray}{0.9}
\usepackage{xspace}
\usepackage{tcolorbox}

\newcounter{messagecounter}
\newcommand{\messagecounter}{\arabic{messagecounter}}
\newenvironment{messagebox}[1]{\begin{tcolorbox}[colback=black!15!white,colframe=white]\refstepcounter{messagecounter}{\textbf{Verdict \messagecounter:}}\label{#1}}{\end{tcolorbox}}

\newif\ifanonymous

\title{\LARGE \bf
Learning and Optimization with 3D Orientations
}

\ifanonymous
\author{Anonymous Authors$^1$%
\thanks{$^{1}$ Anonymous Affiliation}}
\else
\author{Alexandros Ntagkas$^{2,3}$, Constantinos Tsakonas$^{1\ddagger}$, Chairi Kiourt$^{3,4}$ and Konstantinos Chatzilygeroudis$^{1,2,3}$
\thanks{*This work was supported by the Hellenic Foundation for Research and Innovation (H.F.R.I.) under the ``3rd Call for H.F.R.I. Research Projects to support Post-Doctoral Researchers'' (Project Acronym: NOSALRO, Project Number: 7541). This work has also been partially supported by project MIS 5154714 of the National Recovery and Resilience Plan Greece 2.0 funded by the European Union under the NextGenerationEU Program.}
\thanks{$^{1}$Computational Intelligence Laboratory (CILab), Department of Mathematics,
        University of Patras, GR-26110 Patras, Greece.}%
\thanks{$^{2}$Laboratory of Automation and Robotics (LAR) in the Department of Electrical \& Computer Engineering,
        University of Patras, GR-26504 Patras, Greece,
        {\tt\small a\_ntagkas@ac.upatras.gr, costashatz@upatras.gr}}%
\thanks{$^{3}$Archimedes/Athena RC, Greece}%
\thanks{$^{4}$Athena - Research and Innovation Center in Information, Communication and Knowledge Technologies, Xanthi, Greece,
        {\tt\small chairiq@athenarc.gr}}%
\thanks{$^{\ddagger}$C. Tsakonas is now with Inria Centre at Université de Lorraine - work done while at CILab, {\tt\small konstantinos.tsakonas@inria.fr}}%
}
\fi

\begin{document}
\maketitle
\thispagestyle{empty}
\pagestyle{empty}

\begin{abstract}
There exist numerous ways of representing 3D orientations. Each representation has both limitations and unique features. Choosing the best representation for one task is often a difficult chore, and there exist conflicting opinions on which representation is better suited for a set of family of tasks. Even worse, when dealing with scenarios where we need to learn or optimize functions with orientations as inputs and/or outputs, the set of possibilities (representations, loss functions, etc.) is even larger and it is not easy to decide what is best for each scenario. In this paper, we attempt to a) present clearly, concisely and with unified notation all available representations, and ``tricks'' related to 3D orientations (including Lie Group algebra), and b) benchmark them in representative scenarios. The first part feels like it is missing from the robotics literature as one has to read many different textbooks and papers in order have a concise and clear understanding of all possibilities, while the benchmark is necessary in order to come up with recommendations based on empirical evidence. More precisely, we experiment with the following settings that attempt to cover most widely used scenarios in robotics: 1) direct optimization, 2) imitation/supervised learning with a neural network controller, 3) reinforcement learning, and 4) trajectory optimization using differential dynamic programming. We finally provide guidelines depending on the scenario, and make available a reference implementation of all the orientation math described.
\end{abstract}

\vspace{-0.5em}
\section{Introduction \& Motivation}\label{sec:intro}
One of the aspects required to fully describe a robot in a three-dimensional space is its orientation. Consequently, tackling challenging and solving novel tasks requires manipulating orientations, or using them as inputs to policies or controllers~\cite{zhou2019on}. However, we do not have a clear picture of what orientation representation is optimal for the task we desire to solve since one has to go through many different textbooks (and often with very different notations, mindsets and target groups) to fully grasp the essence of the orientation representations~\cite{huang2019generalized}.

In this paper, we attempt to address this challenge, and we hope that we will help roboticists to have a deeper understanding of the various orientation representations available while also giving mild recommendations of which is the optimal representation for the challenges they tackle. Firstly, we aspire to offer a clear and concise exposition of the various representations available for 3D orientations such as Euler angles, Axis-Angle,  and Quaternions, as well as working in directly with the Special Orthogonal Group and its tangent space (Lie Algebra). This endeavor seeks to fill a notable gap in the robotics literature, where a cohesive understanding of all available options often requires traversing multiple disparate sources.

We, also, present a comparison between these representations in \textbf{representative scenarios}, covering the most common tasks encountered in robotics. Our goal is to provide empirical evidence to guide practitioners in selecting the most appropriate representation for their specific applications. Our targeted scenarios consist of direct orientation optimization, supervised learning and imitation learning with neural networks, reinforcement learning, trajectory optimization using differential dynamic programming. We are confident that the insights from solving these tasks will apply to a broad spectrum of robotic tasks.

Finally, we provide a reference Python implementation of all the orientation mathematics described in this paper\footnote{Available at \url{https://github.com/nosalro/orientation_math}.}. This implementation includes comprehensive functions and utilities to handle the various operations discussed, such as conversions between different orientation representations, interpolation techniques for smooth transitions, and analytic gradients for all important operations. By making this implementation available, we aim to facilitate reproducibility, enable deeper understanding of the concepts, and assist researchers and practitioners in applying the theoretical concepts to practical problems.

We advocate that the contribution of our work will benefit the community, not only by acquiring insights from the different benchmarks but also by having a go-to source when working with orientations in a three-dimensional space. We hope that our work will help the decision-making process for practitioners struggling with the complexities of 3D orientations and will foster a deeper understanding of the nuances inherent in representation selection.

\vspace{-0.5em}
\section{Background}\label{sec:bg}
%
%
Before delving deeper into the different orientation representations, we first need to define some fundamental aspects, essential for building an intuition about and working with orientations. In a 3D space, every rotation is fully described by a transformation matrix called the \emph{rotation matrix}. For any rotation occurring about one of the principal axes, we can obtain the corresponding rotation matrix as follows:
\begin{equation}
\label{eq:Rx}
    \boldsymbol{R}_X(\phi) = 
    \begin{bmatrix}
    1 & 0 & 0 \\
    0 & \cos{\phi} & -\sin{\phi} \\
    0 & \sin{\phi} & \cos{\phi}
    \end{bmatrix}
\end{equation}
\begin{equation}
\label{eq:Ry}
    \boldsymbol{R}_Y(\theta) = 
    \begin{bmatrix}
    \cos{\theta} & 0 & \sin{\theta} \\
    0 & 1 & 0 \\
    -\sin{\theta} & 0 & \cos{\theta}
    \end{bmatrix}
\end{equation}
\begin{equation}
\label{eq:Rz}
    \boldsymbol{R}_Z(\psi) = 
    \begin{bmatrix}
    \cos{\psi} & -\sin{\psi} & 0\\
    \sin{\psi} & \cos{\psi} & 0\\
    0 & 0 & 1
    \end{bmatrix}
\end{equation}
Nevertheless, rotation matrices differ from the vector quantities we are familiar with. A set of rotations is not a vectorspace but a non-commutative group, which shares some, but not all, of the properties of a vectorspace~\cite{barfoot2024state}.
A useful operator that will ``pop-up'' in many cases is the skew-symmetric operator:
\begin{equation}
    \boldsymbol{\phi}^\land = \begin{bmatrix}\phi_1\\\phi_2\\\phi_3\end{bmatrix}^\land = \begin{bmatrix}0 & -\phi_3 & \phi_2\\\phi_3 & 0 & -\phi_1\\-\phi_2 & \phi_1 & 0\end{bmatrix}
\end{equation}
We define $(\cdot)^\lor$ as the inverse of the skew-symmetric operator.
\subsection{Special Orthogonal Group}\label{sec:so3}
The special orthogonal group ($SO(3)$) is a group that includes all feasible rotations, so that: 
\begin{equation}
    \label{eq:SO3}
    SO(3) = \{ \boldsymbol{R} \in \mathbb{R}^{3x3} \; | \ \boldsymbol{R}\boldsymbol{R}^T = \boldsymbol{I},\ \text{det}(\boldsymbol{R}) = 1\}
\end{equation}
where $\boldsymbol{I}$ is the identity matrix.
$SO(3)$ is not a vectorspace, and this is based upon two key characteristics of the group. Initially, a zero matrix is not a valid rotation: $\textbf{0}\notin SO(3)$. Secondly, $SO(3)$ is not closed under the addition operation: 
\begin{equation}
    \label{eq:SO3Add}
    \boldsymbol{R}_1, \boldsymbol{R}_2 \in SO(3) \centernot\implies \boldsymbol{R}_1 + \boldsymbol{R}_2 \in SO(3)
\end{equation}
The lack of these properties prevents $SO(3)$ from being a vectorspace. $SO(3)$, however, is a matrix Lie group, and thus enables us to perform useful analysis~\cite{barfoot2024state,sola2018micro}. A Lie group is a group in which the operations are smooth, meaning that we can apply differential calculus on its elements, and it is also a differentiable manifold. The combination of elements belonging to a matrix Lie group can be done by multiplication, whereas the matrix inversion is the inversion operation of the group. The properties of the matrix Lie group (for $SO(3)$) are presented in Table~\ref{tab:prop_lie_group}.

\begin{table}[h]
    \centering
    \begin{tabular}{c c}
    Property & $SO(3)$ \\
    \hline \\
    Closure    &  $\boldsymbol{R}_1, \boldsymbol{R}_2 \in SO(3) \implies \boldsymbol{R}_1\boldsymbol{R}_2 \in SO(3)$ \\
    Associativity    &  $\boldsymbol{R}_1(\boldsymbol{R}_2\boldsymbol{R}_3) = (\boldsymbol{R}_1\boldsymbol{R}_2)\boldsymbol{R}_3 = \boldsymbol{R}_1\boldsymbol{R}_2\boldsymbol{R}_3$ \\
    Identity    &  $\boldsymbol{R}\boldsymbol{I} = \boldsymbol{I}\boldsymbol{R} = \boldsymbol{R}$ \\
    Invertibility    &  $\boldsymbol{R}^{-1}\in SO(3)$
    \end{tabular}
    \caption{Properties of the $SO(3)$ matrix Lie group}
    \label{tab:prop_lie_group}
    \vspace{-2em}
\end{table}
%
%
\subsubsection{Lie Algebra}\label{sec:liealg}
Every matrix Lie group comes with an algebra associated with the group, called Lie algebra. The Lie algebra is defined by a vecotrspace, $\mathbb{V}$, over some field, $\mathbb{F}$, along with a binary operator $[\cdot, \cdot]$, called the Lie bracket~\cite{barfoot2024state}. 
%
%
The Lie algebra associated with $SO(3)$ is defined as follows:
\begin{align}
    &\text{Vectorspace:}\ \mathfrak{so}(3) = \{\boldsymbol{\Phi} = \boldsymbol{\phi}^\land \in \mathbb{R}^{3x3} | \boldsymbol{\phi} \in \mathbb{R}^3\},\nonumber\\
    &\text{Field:} \ \mathbb{R},\nonumber\\
    &\text{Lie bracket:}\ [\boldsymbol{\Phi}_1, \boldsymbol{\Phi}_2] = \boldsymbol{\Phi}_1\boldsymbol{\Phi}_2 - \boldsymbol{\Phi}_2\boldsymbol{\Phi}_1,
\end{align}

\subsubsection{Exponential and Logarithmic Maps}

The exponential map, \(\exp: \mathfrak{so}(3) \rightarrow SO(3)\), connects the Lie algebra to the Lie group. For a given \(\boldsymbol{\phi}\in\mathfrak{so}(3)\), the exponential map is defined as:
 \begin{align}
     \boldsymbol{R} &=\exp(\boldsymbol{\phi})\nonumber\\
     &=\cos(\theta)I+(1-\cos(\theta))\boldsymbol{a}\boldsymbol{a}^T+\sin(\theta)\boldsymbol{a}^\land
\end{align}
where $\boldsymbol{\phi}=\theta\boldsymbol{a}$.
This map is crucial for integrating infinitesimal rotations to obtain finite rotations. Conversely, the logarithmic map, \(\log: SO(3)\rightarrow\mathfrak{so}(3)\), allows us to retrieve the Lie algebra element from a rotation matrix\footnote{We do not include handling a few edge cases for clarity.}:
\begin{align}
    \boldsymbol{\phi} = \log(\boldsymbol{R})=\frac{\theta}{2\sin(\theta)}(\boldsymbol{R}-\boldsymbol{R}^T)^\lor
\end{align}
where $\theta=\cos^{-1}(\frac{Tr(\boldsymbol{R})-1}{2})$.

These maps are essential for moving between the group and its algebra, facilitating operations such as interpolation and differentiation on \(SO(3)\).

\subsubsection{Plus and Minus Operators}

The \(\oplus\) and \(\ominus\) operators are used to represent the addition and subtraction of elements in the tangent space of \(SO(3)\). These operators are defined as follows:

- \textbf{Right \(\oplus\) operation}:
  \begin{align}
  \boldsymbol{R}_{k+1} = \boldsymbol{R}_k \oplus \delta \phi = \boldsymbol{R}_k \exp(\delta\boldsymbol{\phi})
  \end{align}

  Here, \(\delta\boldsymbol{\phi}\) belongs to the tangent space of \(\boldsymbol{R}_k\).

- \textbf{Left \(\oplus\) operation}:
\begin{align}
  \boldsymbol{R}_{k+1} = \delta\boldsymbol{\phi} \oplus \boldsymbol{R}_k = \exp(\delta\boldsymbol{\phi})\boldsymbol{R}_k
\end{align}
  In this case, \(\delta\boldsymbol{\phi}\) belongs to the tangent space of the identity.

Similarly, the \(\ominus\) operator is used to compute the difference between two rotation matrices:

- \textbf{Right \(\ominus\) operation}:
\begin{align}
     \delta\boldsymbol{\phi} = \boldsymbol{R}_{k+1} \ominus \boldsymbol{R}_k = \log(\boldsymbol{R}_k^T\boldsymbol{R}_{k+1})
\end{align}

- \textbf{Left \(\ominus\) operation}:
\begin{align}
      \delta\boldsymbol{\phi} = \boldsymbol{R}_{k+1} \ominus \boldsymbol{R}_k = \log(\boldsymbol{R}_{k+1} \boldsymbol{R}_k^T)
\end{align}

These operators are fundamental for performing updates in optimization and estimation problems on \(SO(3)\).
\subsubsection{Adjoint Matrix}
The Adjoint Matrix, \(\text{Ad}_{\boldsymbol{R}}\), plays a significant role in transforming elements of the Lie algebra under the action of the group~\cite{sola2018micro}. For \(SO(3)\), the adjoint operator is simply the rotation matrix itself:
\begin{align}
\text{Ad}_{\boldsymbol{R}} = \boldsymbol{R}
\end{align}

This operator has several important properties:
\begin{align}
\text{Ad}_{\boldsymbol{R}^{-1}} = \text{Ad}_{\boldsymbol{R}}^{-1}&, \quad \text{Ad}_{\boldsymbol{R}_a\boldsymbol{R}_b} = \text{Ad}_{\boldsymbol{R}_a}\text{Ad}_{\boldsymbol{R}_b},\nonumber\\
\boldsymbol{R} \oplus \delta\boldsymbol{\phi} &= (\text{Ad}_{\boldsymbol{R}}\delta\boldsymbol{\phi}) \oplus \boldsymbol{R}
\end{align}
%
%
\subsubsection{Derivatives on the Manifold}\label{sec:so3_derivs}
When working with functions defined on $SO(3)$, derivatives are computed using the tangent space and the $\oplus$ and $\ominus$ operators. The Jacobian of a function $f: \mathbb{R}^N \to \mathbb{R}^M$ is typically defined as:

\begin{equation*}
\boldsymbol{J} = \frac{\partial f(\boldsymbol{x})}{\partial\boldsymbol{x}} = \lim_{\delta\boldsymbol{x}\to 0}\frac{f(\boldsymbol{x} + \delta\boldsymbol{x})-f(\boldsymbol{x})}{\delta\boldsymbol{x}}
\end{equation*}

For functions involving \(SO(3)\), we extend this definition using the right \(\oplus\) and \(\ominus\) operators:
\begin{equation*}
\boldsymbol{J} = \frac{\prescript{\boldsymbol{R}}{}D f(\boldsymbol{R})}{\partial\boldsymbol{R}} = \lim_{\delta\boldsymbol{\phi}\to 0}\frac{f(\boldsymbol{R}\oplus\delta\boldsymbol{\phi})\ominus f(\boldsymbol{R})}{\delta\boldsymbol{\phi}}
\end{equation*}

This is known as the \textbf{right Jacobian}. It can also be expressed in terms of the logarithmic map:
\begin{equation*}
\boldsymbol{J} = \frac{\partial \log(f(\boldsymbol{R})^{-1}f(\boldsymbol{R})\exp(\delta\boldsymbol{\phi}^\land))}{\partial\delta\boldsymbol{\phi}}\Big\vert_{\delta\boldsymbol{\phi}=0}
\end{equation*}

Similarly, using the left \(\oplus\) and \(\ominus\) operators, we define the \textbf{left Jacobian}:

\begin{equation*}
\boldsymbol{J} = \frac{\prescript{\boldsymbol{I}}{}D f(\boldsymbol{R})}{\partial\boldsymbol{R}} = \lim_{\delta\boldsymbol{\phi}\to 0}\frac{f(\delta\boldsymbol{\phi}\oplus\boldsymbol{R})\ominus f(\boldsymbol{R})}{\delta\boldsymbol{\phi}}
\end{equation*}
This can also be written as:
\[
\boldsymbol{J} = \frac{\partial \log(\exp(\delta\boldsymbol{\phi}^\land)f(\boldsymbol{R})f(\boldsymbol{R})^{-1})}{\partial\delta\boldsymbol{\phi}}\Big\vert_{\delta\boldsymbol{\phi}=0}
\]

Unless otherwise specified, the \textbf{right Jacobian} is typically used for differentiation on \(SO(3)\).
The chain rule applies to functions on \(SO(3)\). For example, if \(\boldsymbol{R}_{k+1} = f(\boldsymbol{R}_k)\) and \(\boldsymbol{R}_{k+2} = g(f(\boldsymbol{R}_k))\), the derivative is given by:

\begin{equation*}
\frac{Dg}{D\boldsymbol{R}_k} = \frac{Dg}{D\boldsymbol{R}_{k+1}}\frac{Df}{D\boldsymbol{R}_k} \quad \text{or} \quad \boldsymbol{J}^g_{\boldsymbol{R}_k} = \boldsymbol{J}^g_{\boldsymbol{R}_{k+1}}\boldsymbol{J}^{f}_{\boldsymbol{R}_k}
\end{equation*}

This allows us to compute derivatives of composite functions on \(SO(3)\) efficiently.
\subsection{Euler Angles}
Perhaps the simplest representation is the Euler angles representation. This representation comprises a set of three angles, which describe the orientation of a rigid body in 3D space. It is worth noting that this representation is widely used in the field of robotics and robot learning due to its simplicity and intuitive interpretation. A set of angles $\boldsymbol{\theta} = [\theta_1, \theta_2, \theta_3] \in \mathbb{R}^3$, which correspond to the rotation of the object about some $X, Y, Z$ axes. The set $\boldsymbol{\theta}$ translates to, firstly, rotating the body about its own Z-axis by the $\theta_3$, then about its new Y axis by $\theta_2$, and finally about the resulting X-axis by $\theta_1$.
There are multiple conventions regarding the Euler angles representation, depending on the application and the requirements of the problem, such as $ZYX$, $ZXZ$, etc. Each of these conventions indicates the order, in which the rotations are applied to the object about the specified axis.
The most common conventions are $XYZ$ (roll-pitch-yaw) and $ZYX$ (yaw-pitch-roll) conventions (rotation around the principal axes). In any case, for an arbitrary $\alpha_1\alpha_2\alpha_3$ convention we can obtain the rotation matrix as follows:
\begin{equation}
    \label{eq:Reuler}
    \boldsymbol{R}_{\alpha_1\alpha_2\alpha_3}(\boldsymbol{\theta}) = \boldsymbol{R}_{\alpha_1}(\theta_1)\boldsymbol{R}_{\alpha_2}(\theta_2)\boldsymbol{R}_{\alpha_3}(\theta_3)
\end{equation}
We can also easily go from a rotation matrix to the parameters $\boldsymbol{\theta}$ (see e.g.~\cite{schneider2002geometric,kindrcheatsheet}).
%

\subsection{Axis-Angle}
Another popular representation is the Axis-Angle. This representation is essentially the same as the tangent space of $SO(3)$.
We usually write $\boldsymbol{\phi} \in \mathbb{R}^3$ as $\boldsymbol{\phi} = \phi\boldsymbol{a}$.
%
%
\subsection{Quaternions}
Quaternions are a mathematical structure that extends complex numbers. Since we are interested in Quaternions as a tool for representing rotations, we actually refer to unit Quaternions ($||\boldsymbol{q}||_2 = 1$). A quaternion is defined as:
%
\begin{equation}
    \label{eq:quaternion_def}
    \boldsymbol{q}\in \mathbb{S} := [q_s\;\; \boldsymbol{q}_v^T]^T
\end{equation}
where $q_s \in \mathbb{R}$ and $\boldsymbol{q}_v \in \mathbb{R}^3$, and $\mathbb{S}$ is the space of unit Quaternions. The multiplication operation between Quaternions can be defined as~\cite{jackson2021planning}: 
\begin{equation}
    \boldsymbol{q}_2 \otimes \boldsymbol{q}_1 = L(\boldsymbol{q}_2)\boldsymbol{q}_1 = R(\boldsymbol{q}_1)\boldsymbol{q}_2
\end{equation}
where $L(\boldsymbol{q})$ and $R(\boldsymbol{q})$ are defined as:
\begin{equation*}
\begin{split}
    L(\boldsymbol{q}) = \begin{bmatrix}
        q_s & -\boldsymbol{q}_v^T \\
        \boldsymbol{q}_v & q_s\boldsymbol{I} + \boldsymbol{q}_v^\land
    \end{bmatrix} \\
    R(\boldsymbol{q}) = \begin{bmatrix}
        q_s & -\boldsymbol{q}_v^T \\
        \boldsymbol{q}_v & q_s\boldsymbol{I} - \boldsymbol{q}_v^\land
    \end{bmatrix},
\end{split}
\end{equation*}
and $L(\boldsymbol{q})$ and $R(\boldsymbol{q})$ are orthonormal matrices.
Some interesting properties ($\boldsymbol{r}\in \mathbb{R}^3$):
\begin{equation}
    \boldsymbol{H}\boldsymbol{r} := \begin{bmatrix}
        0 \\
        \boldsymbol{I}
    \end{bmatrix}\boldsymbol{r},
   \boldsymbol{q}^* = \boldsymbol{T}\boldsymbol{q} := \begin{bmatrix}
       1 & \  \\
       \ & -\boldsymbol{I}
   \end{bmatrix}\boldsymbol{q}
\end{equation}
It is worth mentioning that the inverse quaternion is equal to $\boldsymbol{q}^*$, hence $\boldsymbol{q}^{-1} = \boldsymbol{q}^*$.
%
When working with Quaternions, rotations are performed through the operation:
\begin{equation}
    \boldsymbol{r}' = \boldsymbol{q} \otimes \boldsymbol{r} \otimes \boldsymbol{q}^* = \boldsymbol{H}^TL(\boldsymbol{q})R(\boldsymbol{q})^T\boldsymbol{H}\boldsymbol{r} = A(\boldsymbol{q})\boldsymbol{r}
\end{equation}
where $A(\boldsymbol{q})$ is the rotation matrix of quaternion $\boldsymbol{q}$.

\subsubsection{Optimization with Quaternions}
When dealing with optimization, we are getting steps of the form $\boldsymbol{x}_{k+1} = \boldsymbol{x}_k + \Delta\boldsymbol{x}$. But as we have seen, adding rotations does not make sense, and the same holds for Quaternions. In other words, $\boldsymbol{q}_{k+1} = \boldsymbol{q}_k + \Delta\boldsymbol{q}$ \emph{does not have any physical meaning}. Besides, as have already seen the \emph{difference} between two orientations/rotations, is itself another rotation, and \emph{the correct way of combining rotations is by multiplication}.

Expanding the previous thought, it makes more sense to compute differences as rotations ($\delta\boldsymbol{\phi}\in\mathbb{R}^3$) than normal differences. By doing so, we are left with an issue: our representation is a $4D$ vector (quaternion), and thus any gradient that we can take should be with respect to the parameters. How can I use $\delta\boldsymbol{\phi}$s with my quaternion representation?

Recently, Jackson et al.~\cite{jackson2021planning} have proposed an \emph{attitude Jacobian ``trick''} that lets us \emph{transform} gradients in the $4D$ quaternion space to the rotation space! In essence, if we have a function $J(\boldsymbol{q}): \mathbb{S}\to\mathbb{R}$, we can compute the derivative in the rotation space as:
\begin{equation}
    \label{eq:attitude_jac}
    \nabla J(\boldsymbol{q}) = \frac{\partial J}{\partial\boldsymbol{q}}L(\boldsymbol{q})\boldsymbol{H} = \frac{\partial J}{\partial\boldsymbol{q}}G(\boldsymbol{q})\in\mathbb{R}^{3\times 3}
\end{equation}
and we call $G(\boldsymbol{q})=L(\boldsymbol{q})\boldsymbol{H}\in\mathbb{R}^{4\times 3}$ the \emph{attitude Jacobian}.
\section{Experimental Scenarios}\label{sec:scenarios}
%
We examine the following settings: 1) direct orientation optimization, 2) imitation learning with neural networks, 3) reinforcement learning, 4) trajectory optimization using differential dynamic programming (iLQR with box constraints in particular~\cite{tassa2014control}). Our experiments focus on tasks that heavily rely on orientation to be carried out successfully, resulting in a more confident comparison between representations.

We experiment with the following environments: 1) Quadrotor performing a flip in 3D, 2) a rotating frame in 3D, and 3) a 7-DOF manipulator (Franka Panda).
We simulated a servo-based 7-DOF arm using the RobotDART simulator~\cite{chatzilygeroudis2024robotdart}, while we created a simulator for the quadrotor and the simple frame experiments.

\subsection{Quadrotor Dynamics in SO(3)}
We detail the dynamics of the quadrotor in the $SO(3)$ space for completeness (the rest of the representations follow an analogous formulation). The state of the quadrotor is:
\begin{align}
    \boldsymbol{x} = \begin{bmatrix}
        \boldsymbol{p}\in\mathbb{R}^3,\text{ position in world frame}\\
        \boldsymbol{R} \in\mathbb{SO}(3),\text{ orientation in world frame}\\
        \boldsymbol{v}\in\mathbb{R}^3,\text{ velocity in body frame}\\
        \boldsymbol{\omega}\in\mathbb{R}^3,\text{ angular velocity in body frame}
    \end{bmatrix}
\end{align}
And the dynamics:
\begin{align}
    \dot{\boldsymbol{p}} &= \boldsymbol{R}\boldsymbol{v},\quad\quad&\dot{\boldsymbol{R}} = \boldsymbol{R}\boldsymbol{\omega}^\land\nonumber\\
    \dot{\boldsymbol{v}} &= \frac{1}{m}\boldsymbol{F} - \boldsymbol{\omega}\times\boldsymbol{v},\quad\quad&\dot{\boldsymbol{\omega}} = \mathcal{I}^{-1}\Big(\boldsymbol{\tau} - \boldsymbol{\omega}\times\mathcal{I}\boldsymbol{\omega}\Big)
\end{align}
$m$ is the mass of the body, $\boldsymbol{F}$ and $\boldsymbol{\tau}$ are the forces and torques applied to the body (expressed in body frame), and $\mathcal{I}$ is the Inertia matrix. Euler integration gives:
\begin{align}
    \boldsymbol{v}_{k+1}&=\boldsymbol{v}_k+\dot{\boldsymbol{v}}_kdt,\quad\quad&\boldsymbol{\omega}_{k+1}=\boldsymbol{\omega}_k+\dot{\boldsymbol{\omega}}_kdt\nonumber\\
    \boldsymbol{p}_{k+1}&=\boldsymbol{p}_k+\boldsymbol{v}_{k}dt,\quad\quad&\boldsymbol{R}_{k+1}=\boldsymbol{R}_k\oplus(\boldsymbol{\omega}_{k}dt)
\end{align}

We linearize over the discrete integration step as follows:
\begin{align}
    \frac{\partial f(\boldsymbol{x}_k,\boldsymbol{u}_k)}{\partial\boldsymbol{x}_k}=
    \begin{bmatrix}
    \frac{\partial\boldsymbol{R}_{k+1}}{\partial\boldsymbol{R}_k} & 0 &0 &\frac{\partial\boldsymbol{R}_{k+1}}{\partial\boldsymbol{\omega}_k} \\  
    \frac{\partial\boldsymbol{p}_{k+1}}{\partial\boldsymbol{R}_k} & \frac{\partial\boldsymbol{p}_{k+1}}{\partial\boldsymbol{p}_k} & \frac{\partial Rp_{n+1}}{\partial\boldsymbol{v}_k} & 0 \\  
    \frac{\partial\boldsymbol{v}_{k+1}}{\partial\boldsymbol{R}_k} & 0 & \frac{\partial\boldsymbol{v}_{k+1}}{\partial\boldsymbol{v}_k} &\frac{\partial\boldsymbol{v}_{k+1}}{\partial\boldsymbol{\omega}_k} \\  
    0&0&0&\frac{\partial\boldsymbol{\omega}_{k+1}}{\partial\boldsymbol{\omega}_k} \\  
    \end{bmatrix}
\end{align}
Using the properties discussed in Section \ref{sec:so3} we can derive the derivatives in $SO(3)$ as follows:
\begin{align}
\frac{\partial\boldsymbol{R}_{k+1}}{\partial\boldsymbol{R}_k} & =\exp(\boldsymbol{\omega}^\land_k dt)^T,\frac{\partial\boldsymbol{p}_{k+1}}{\partial\boldsymbol{R}_k}=-\boldsymbol{R}_k\boldsymbol{v}^\land_kdt\nonumber\\ 
\frac{\partial\boldsymbol{R}_{k+1}}{\partial\boldsymbol{\omega}_k} & =J^{\boldsymbol{R}_k\oplus\boldsymbol{\omega}_k dt}_{\boldsymbol{\omega}_kdt}J^{\boldsymbol{\omega}_kdt}_{\boldsymbol{\omega}_k} = J_r(\boldsymbol{\omega}_k dt)dt\nonumber\\ 
\frac{\partial\boldsymbol{v}_{k+1}}{\partial\boldsymbol{R}_k}&=J^{\boldsymbol{R}^T_k\boldsymbol{v}_k}_{\boldsymbol{R}^T_k}J^{\boldsymbol{R}^T_k}_{\boldsymbol{R}_k}dt=\boldsymbol{R}^T_k\boldsymbol{v}^\land_k\boldsymbol{R}_kdt
\end{align}

Using the same properties, we also get the Jacobian $\frac{\partial f(\boldsymbol{x}_k,\boldsymbol{u}_k)}{\partial\boldsymbol{u}_k}$. In our implementation, we perform a \emph{Semi-Implicit Euler Integration} and adapt the Jacobians/linearization accordingly. We used the similar derivations for the simple frame scenario.

\section{Experimental Results}\label{sec:experiments}
\subsection{Direct Optimization}
\begin{figure}[!htb]
    \centering
    \includegraphics[width=0.9\linewidth]{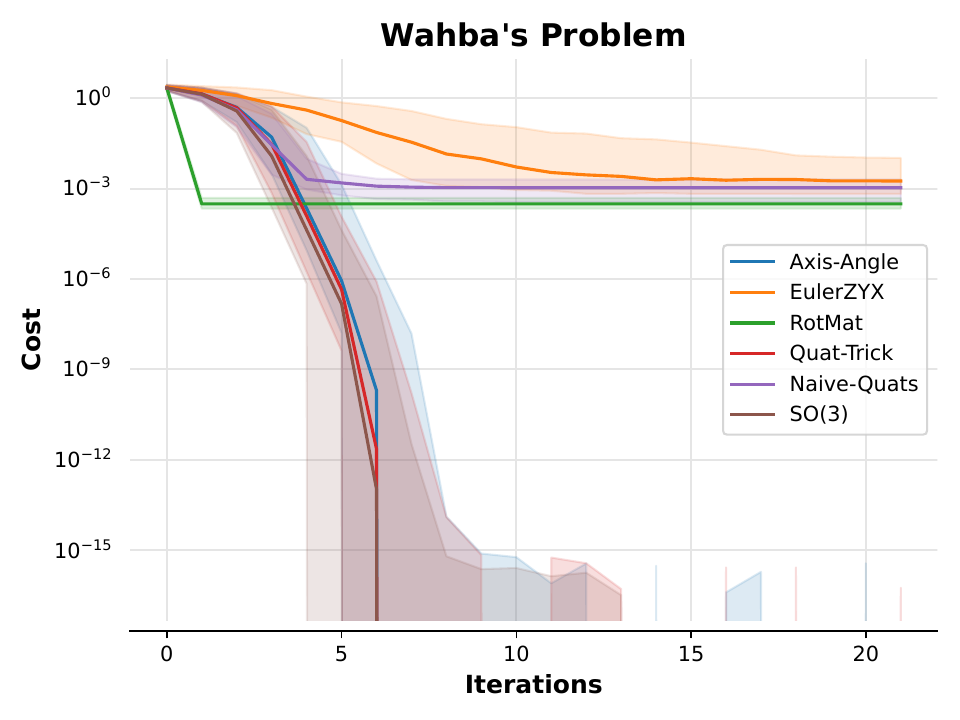}
    \vspace{-1em}
    \caption{Wahba's problem. The plot is in log-scale. Solid lines are the median over 200 replicates and the shaded regions are the regions between the 25-th and 75-th percentiles.}
    \label{fig:wahba}
    \vspace{-1em}
\end{figure}
There are multiple scenarios that directly optimizing for orientations is required. To investigate the best possible representation for such cases, we solve Wahba's Problem~\cite{forbes2015fundamentals,wahba1965least}. This problem focuses on finding the optimal rotation matrix using the least squares method. It can be formulated as follows:
\begin{equation}
    \underset{\boldsymbol{\theta}}{\min} \sum_i ||\boldsymbol{p}^i_w - \boldsymbol{R}_{\mathcal{X}}(\boldsymbol{\theta})\boldsymbol{p}^i_b||^2_2 = ||r(\boldsymbol{\theta})||_2^2
\end{equation}
where $\boldsymbol{p}^i_w, \boldsymbol{p}^i_b$ are the points in world and body frame coordinates respectively. Here, we denote with $\boldsymbol{\theta}$ the parameters of any orientation representation, and $\boldsymbol{R}_{\mathcal{X}}$ is the function that maps a point from an orientation representation $\mathcal{X}$ to the corresponding rotation matrix. Given an iterate $\boldsymbol{\theta}_k$, and using the Gauss-Newton algorithm, we get an update $\delta\boldsymbol{\theta}_k = (\nabla r^T\nabla r)^{-1}\nabla r^Tr(\boldsymbol{\theta}_k)$.
We report the norm of the rotation difference
between iterates and the SVD solution of the problem, and compare the following:
\begin{enumerate}
    \item Rotation Matrices\footnote{We ``flatten'' the rotation matrix, $\boldsymbol{R}\in\mathbb{R}^{3\times 3}$, into $\boldsymbol{\theta}\in\mathbb{R}^9$.}, aka $\boldsymbol{\theta}\in\mathbb{R}^9$;
    \item Euler Angles ZYX, aka $\boldsymbol{\theta}\in\mathbb{R}^3$;
    \item Axis-Angle, aka $\boldsymbol{\theta}\in\mathbb{R}^3$;
    \item Quaternions, aka $\boldsymbol{\theta}\in\mathbb{R}^4$;
    \item Quaternions with the attitude Jacobian (Eq.~\eqref{eq:attitude_jac})~\cite{jackson2021planning};
    \item $SO(3)$ Gauss Newton update~\cite{barfoot2024state}.
\end{enumerate}
The results showcase that the attitude Jacobian ``trick'' and working in $SO(3)$ (or its tangent space, Axis-Angle) clearly outperform the rest of the representations and converge to machine precision accuracy in very few iterations (Fig.~\ref{fig:wahba}). Interestingly flattened rotation matrices work better than naive quaternions.

\begin{messagebox}{msg:direct}
    When directly optimizing for orientations, \emph{we strongly suggest to work directly with the $SO(3)$ space (see Sec.~\ref{sec:so3_derivs})} (or its tangent space).
    ~\\\textbf{Recommended readings:}~\cite{sola2018micro,barfoot2024state}.
\end{messagebox}
\subsection{Imitation Learning}
\begin{figure}[!htb]
    \centering
    \vspace{1em}
    \includegraphics[width=0.6\linewidth]{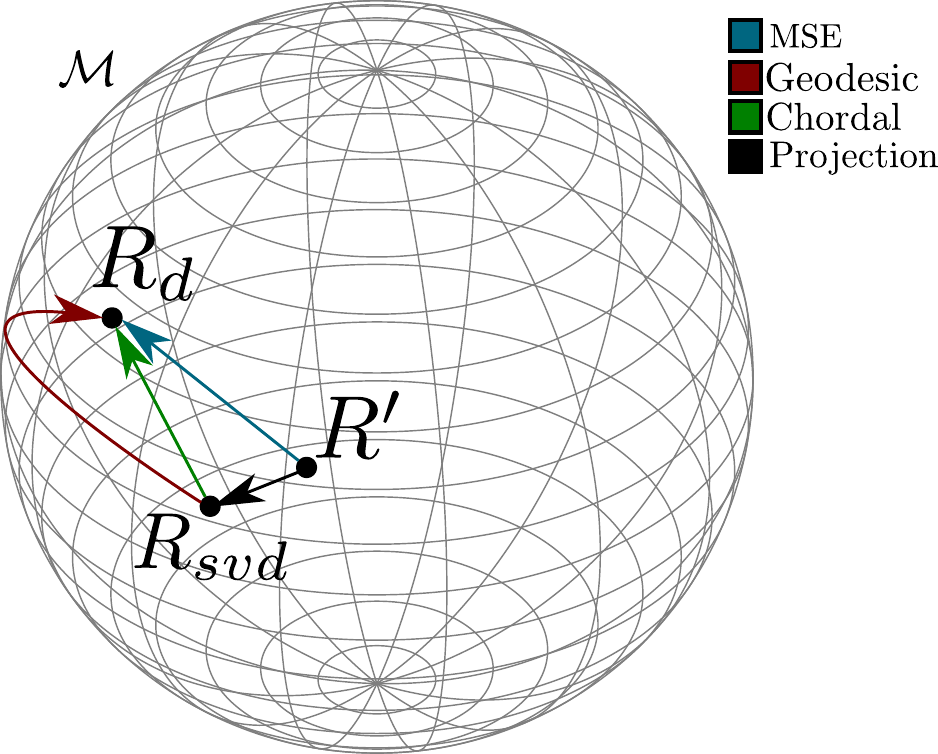}
    \caption{Graphical Explanation of $SO(3)$ losses. Although $SO(3)$ is not a 3D sphere (which is not even a smooth manifold), we use it for visualization.}
    \label{fig:SO3_loss}
    \vspace{-1em}
\end{figure}

%
Imitation learning enables robots to perform complex tasks by mimicking expert demonstrations,
%
In order to isolate the pros and cons of the orientation representations, we devise two experiments: one where we want to rotate a stationary frame to a desired orientation, and one where a robotic arm has to reach a pre-specified 6D position. We collect  demonstrations from random starting orientations using a task space PD controller for the robotic arm and a simple interpolation method for the stationary frame. We train a neural network that gets as input the current orientation and predicts the next close ($k+T$) target to track. We evaluate the performance of the learned controllers on the predefined and random starting orientations to assess the generalization capability each orientation representation offers.
More formally, given a data set
$\mathcal{D} = \{ (\boldsymbol{R}_{k}, \boldsymbol{R}_{k+T}) \}_{i=1}^{N}$
of $N$ inputs, we wish to find the parameters $\boldsymbol{\psi}$ of the neural network $f_{\boldsymbol{\psi}}: \mathcal{X}\to\mathcal{X}$, where $\mathcal{X}$ is one of the orientation representation spaces,
that minimize the loss function
\begin{align}
    \mathcal{L}(\mathcal{D}, \boldsymbol{\psi}) = \frac{1}{N}\sum_{(\boldsymbol{R}_{k}, \boldsymbol{R}_{k+T}) \in \mathcal{D}} d(\boldsymbol{R}_{k+T}, \boldsymbol{R}^{\boldsymbol{\psi}}_{\text{pred}})
\end{align}
using stochastic gradient descent, where $\boldsymbol{R}^{\boldsymbol{\psi}}_{\text{pred}} = \boldsymbol{R}_{\mathcal{X}}(\boldsymbol{\theta}_{\boldsymbol{\psi}})$, $\boldsymbol{\theta}_{\boldsymbol{\psi}} = f_{\boldsymbol{\psi}}(\boldsymbol{\theta}_{\mathcal{X}}(\boldsymbol{R}_{k}))$, $\boldsymbol{R}_{\mathcal{X}}$ is the function that maps a point from an orientation representation $\mathcal{X}$ to the corresponding rotation matrix and $\boldsymbol{\theta}_{\mathcal{X}}$ is the inverse of $\boldsymbol{R}_{\mathcal{X}}$. Both the representation $ \mathcal{X}$ and the distance function $d$ play a crucial role in the learning convergence and policy performance.\\
We experiment with three different distance functions between the desired rotation matrix $\boldsymbol{R}_d$ and the prediction of the neural network $\boldsymbol{R}'$ (Fig.~\ref{fig:SO3_loss}):
\begin{enumerate}
    \item \textbf{Naive:} $d_{mse}(\boldsymbol{R}_d,\boldsymbol{R}')= \|\boldsymbol{R}_d-\boldsymbol{R}'\|_2$
    \item \textbf{Geodesic:} $d_{geo}(\boldsymbol{R}_d,\boldsymbol{R}')= \|\boldsymbol{R}_d\ominus\boldsymbol{R}'\|$
    \item \textbf{Chordal:} $d_{ch}(\boldsymbol{R}_d,\boldsymbol{R}')= \|\boldsymbol{R}_d-\boldsymbol{R}_{svd}\|_2$
\end{enumerate}
where $\boldsymbol{R}_{svd}$ describes the element of $SO(3)$ with the least-squares distance to $\boldsymbol{R}'$, using the analysis in~\cite{svd}. Note that depending on the representation it is possible for $\boldsymbol{R}'$ to not be a valid rotation matrix, since this is a prediction of a neural network. In our case, this happens when we use ``flattened'' rotation matrices as a representation; we use the Chordal distance only for this representation as for the others the Chordal and Naive difference are identical. Another important detail is that we rely on autodiff tools to handle the projection operation, \emph{ensuring its gradient is computed during backpropagation}. Moreover, in this scenario we were not able to use the attitude Jacobian ``trick'' or work directly in the $SO(3)$ space since the optimization variables do not include any orientation representation.

We evaluate the model's performance every epoch and we perform 15 replicates of the task (initial point is randomly selected), and we compare the following representations:
\begin{enumerate}
    \item Rotation Matrices, aka $\boldsymbol{\theta}\in\mathbb{R}^9$,
    \item Euler Angles ZYX, aka $\boldsymbol{\theta}\in\mathbb{R}^3$;
    \item Axis-Angle, aka $\boldsymbol{\theta}\in\mathbb{R}^3$;
    \item Quaternions, aka $\boldsymbol{\theta}\in\mathbb{R}^4$;
\end{enumerate}

\begin{figure}[!htb]
    \centering
    \includegraphics[width=\linewidth]{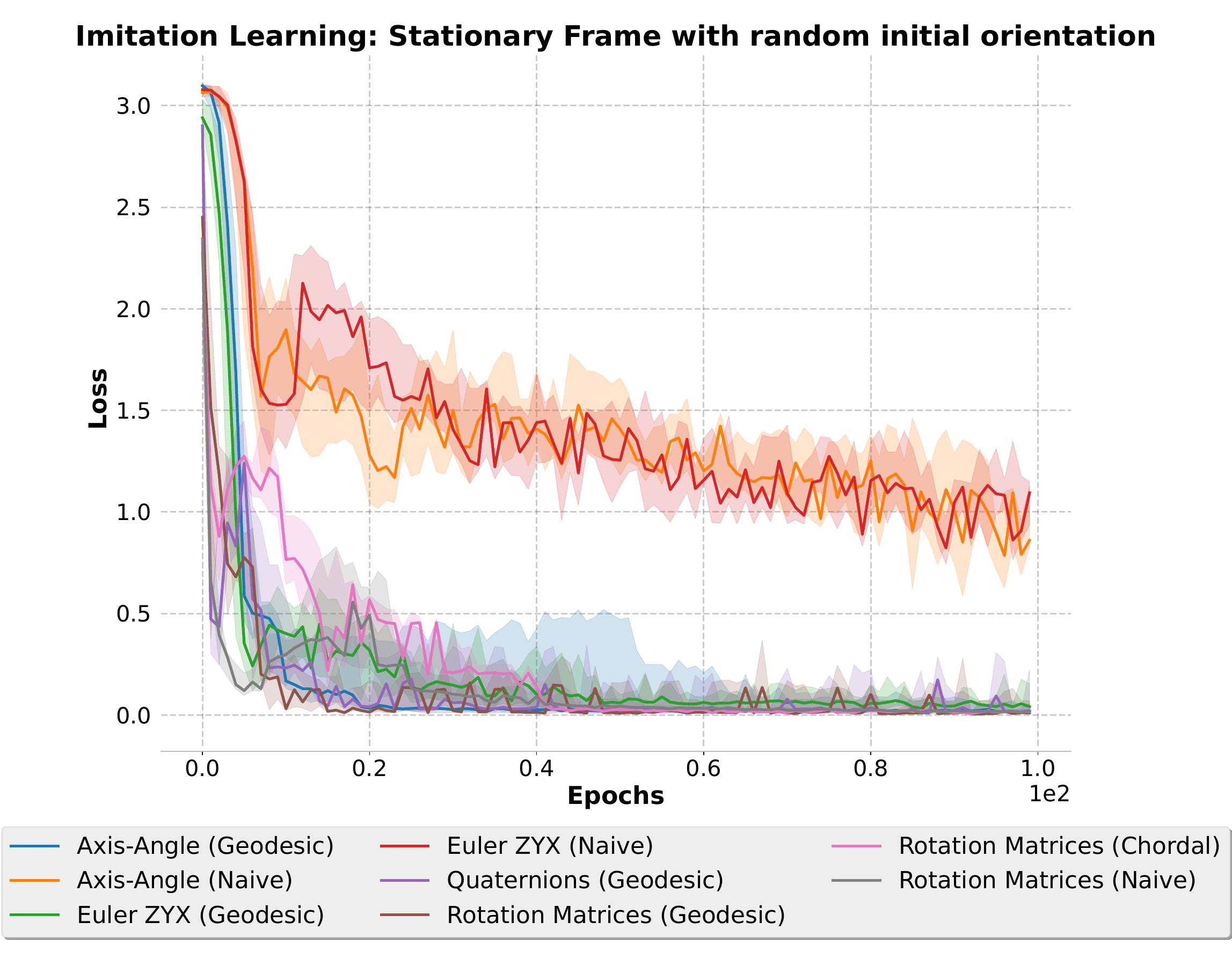}
    \caption{Imitation Learning: Stationary frame reaching desired orientation
    through learning from demonstrations. Solid lines are the median over 10
    replicates and the shaded regions are the regions between the 25-th and 75-
    th percentiles.}
    \label{fig:imitation_fixed}
    \vspace{-1em}
\end{figure}
\begin{figure}[!htb]
    \centering
    \includegraphics[width=\linewidth]{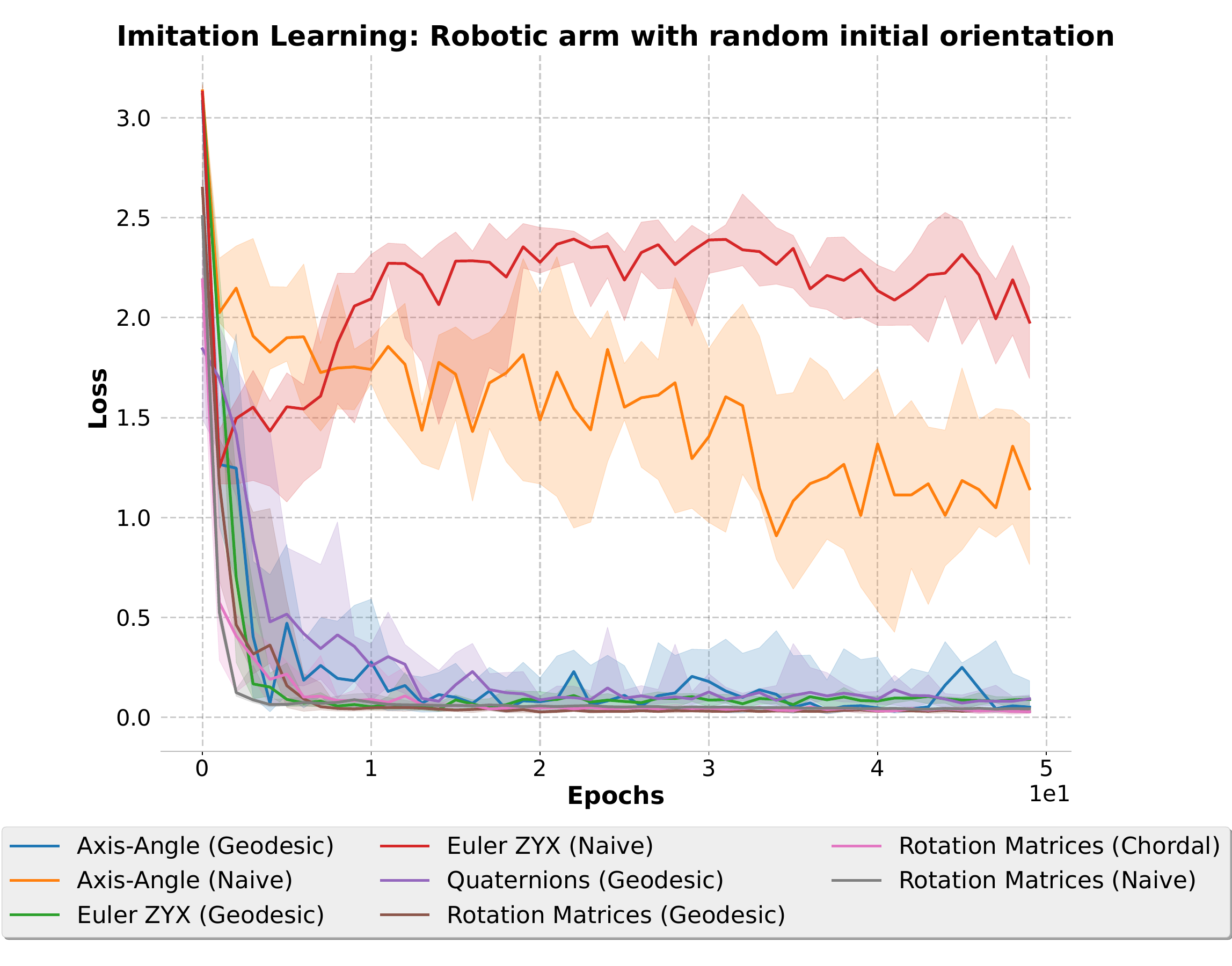}
    \caption{Imitation Learning: Robotic arm reaching desired orientation and
    position through learning from demonstrations. Solid lines are the median
    over 5 replicates and the shaded regions are the regions between the 25-th
    and 75-th percentiles.}
    \label{fig:imitation_random}
    \vspace{-1em}
\end{figure}

The results showcase that encoding orientations as rotation matrices and using the \emph{Chordal} distance has the best performance overall, while Quaternions and Axis-Angle with the \emph{Geodesic} distance have similar results (Fig.~\ref{fig:imitation_fixed},~\ref{fig:imitation_random}). While this might seem counter-intuitive, our results coincide with recent results in the machine learning literature~\cite{geist2024learning}. We, also, observe that when using the Euler Angles and Axis-Angle representations with the naive distance we cannot learn reliable targets, while both representations seem to achieve better results when using the geodesic distance.
\begin{messagebox}{msg:imitation}
    In imitation learning settings, we suggest to use \emph{``flattened Rotation Matrices''} representation along with the \textbf{Chordal} distance. Quaternions and Axis-Angle with the Geodesic distance also provide effective learning and generalization. Overall, it appears that the loss function is more important than the input/output representation.
    ~\\\textbf{Recommended readings:}~\cite{geist2024learning}.
\end{messagebox}
\subsection{Reinforcement Learning}
In order to identify which representation is crucial as input to RL policies, we devise the following experiments. The first one will be a stationary frame that has to reach a desired orientation, and the second one is a 3D quadrotor needs to reach a target point while performing a flip in the process. The simulator is always using the $SO(3)$ integration and we adapt the input to the policy according to the representation.
In order to learn the task, we use the Proximal Policy Optimization (PPO) algorithm~\cite{schulman2017proximal} and the reward function:
\begin{equation}
    \label{eq:rl_reward}
    r(\boldsymbol{s}_t) = ||\boldsymbol{p}_g - \boldsymbol{p}_t||_2^2 + |\boldsymbol{\theta}_{gt}\ominus \boldsymbol{\theta}_{t}|,
\end{equation}
where $\boldsymbol{p}_g, \boldsymbol{\theta}_{gt}$\footnote{We have defined an orientation ``flip'' trajectory.} are the target point and orientation target at time $t$ respectively, and $\boldsymbol{p}_t, \boldsymbol{\theta}_{t}$ the position and orientation of the quadrotor at time $t$. The first component ensures that our quadrotor will end up at the target, and the second one makes the robot to follow the desired orientation trajectory.
%
%
\begin{figure}[!htb]
    \centering
    \includegraphics[width=\linewidth]{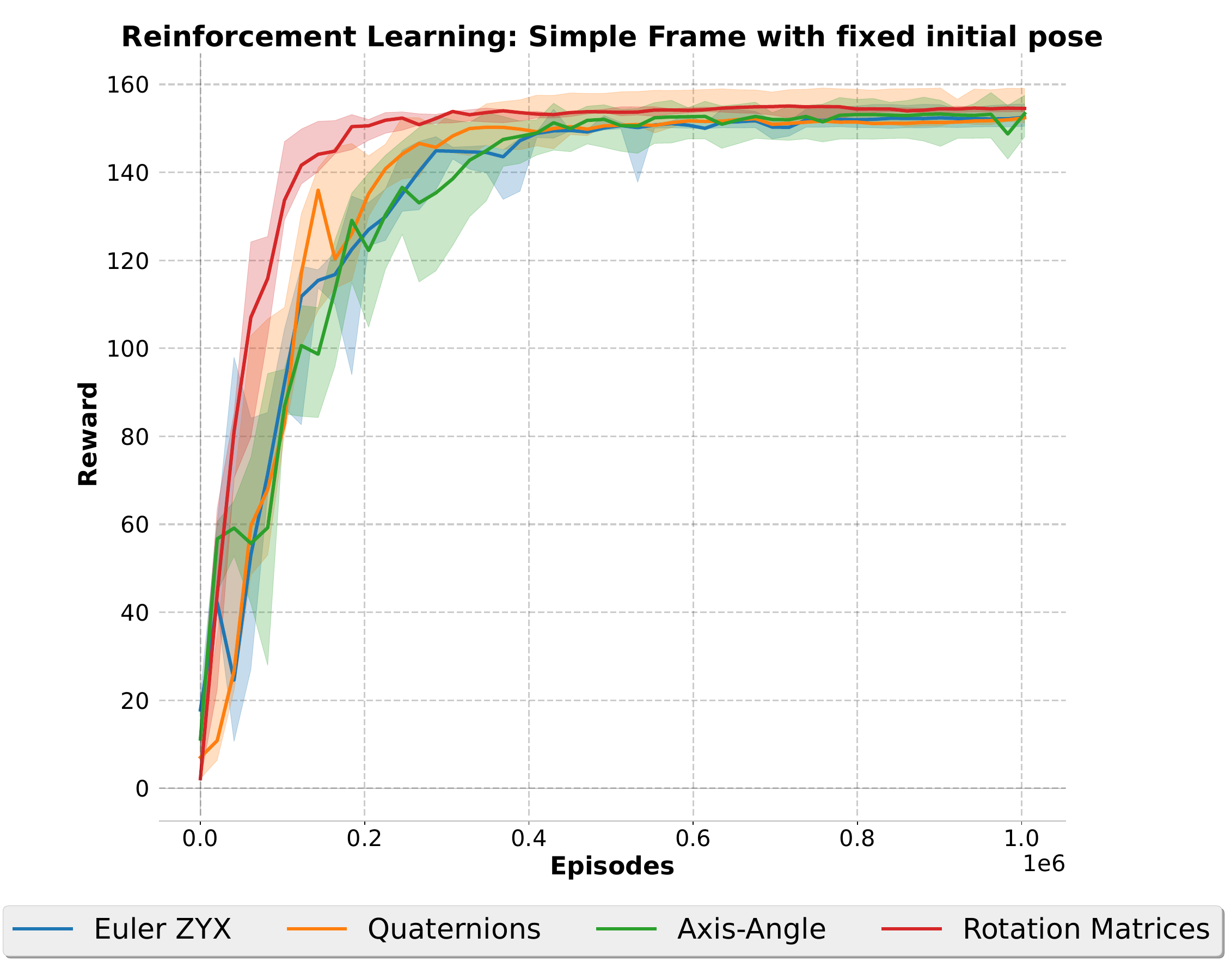}
    \caption{Frame Alignment: During training and evaluation, the frame starts from a predefined position. Solid lines are the median over 5 replicates and the shaded regions are the regions between the 25-th and 75-th percentiles.}
    \label{fig:rl_frame_fixed}
    \vspace{-1em}
\end{figure}
\begin{figure}[!htb]
    \centering
    \includegraphics[width=\linewidth]{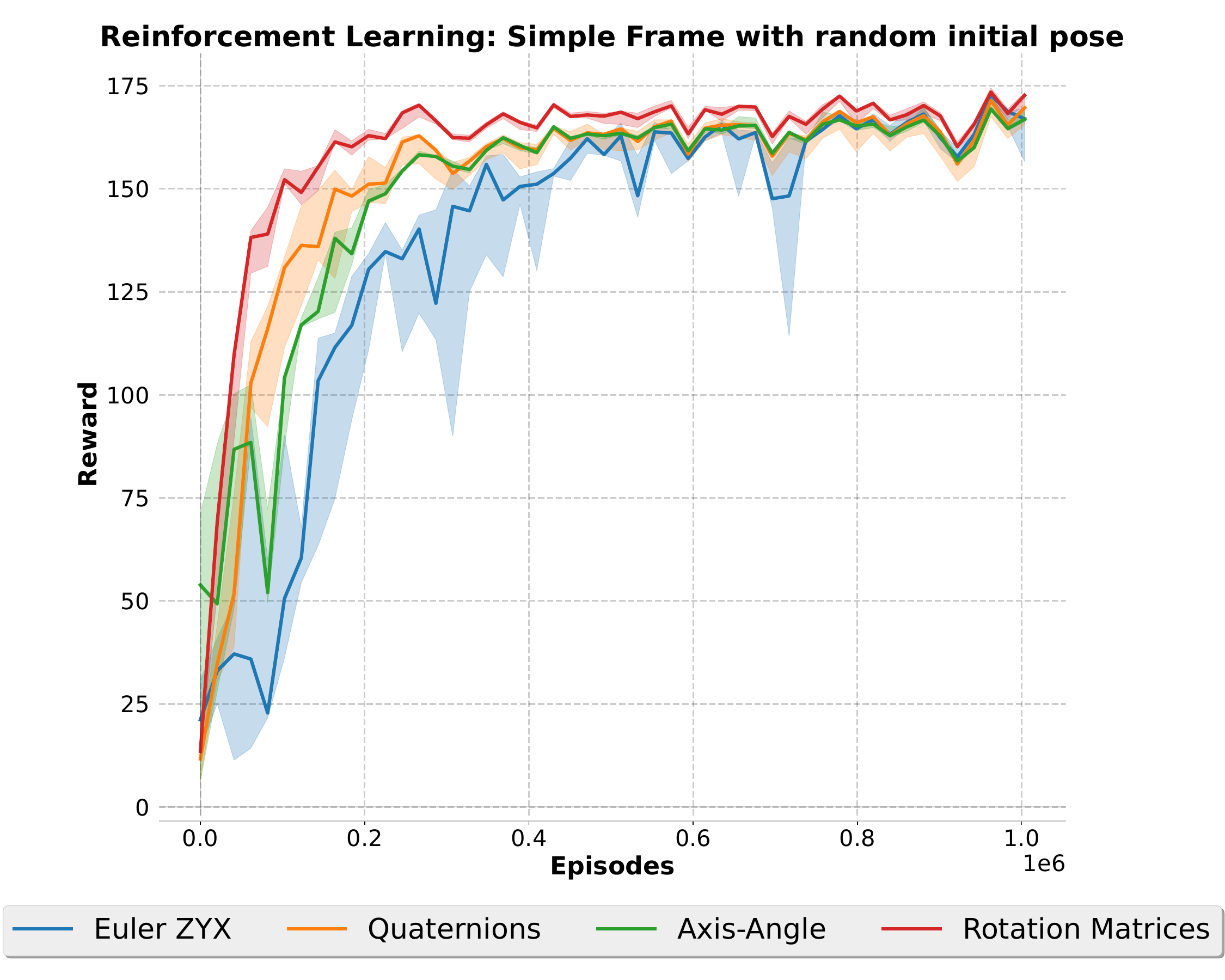}
    \caption{Frame Alignment: During training and evaluation, the frame starts from a  random position. Solid lines are the median over 5 replicates and the shaded regions are the regions between the 25-th and 75-th percentiles.}
    \label{fig:rl_frame_random}
    \vspace{-1em}
\end{figure}
%
%
%
%

%
%
\begin{figure}[!htb]
    \centering
    \includegraphics[width=\linewidth]{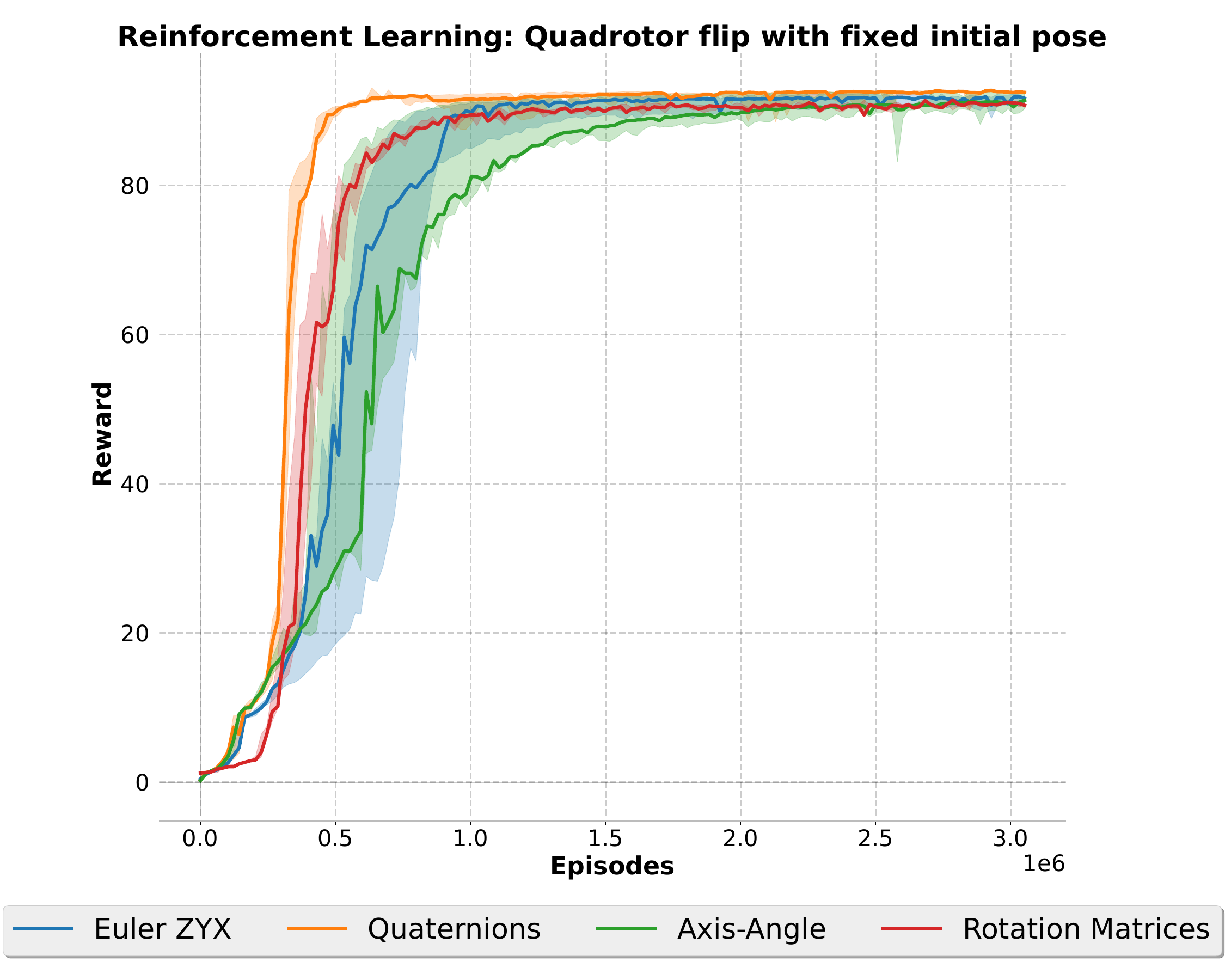}
    \caption{Quadrotor Flip RL Task: During training and evaluation, the robot starts from a predefined position. Solid lines are the median over 5 replicates and the shaded regions are the regions between the 25-th and 75-th percentiles.}
    \label{fig:rl_quad_fixed}
    \vspace{-1em}
\end{figure}
\begin{figure}[!htb]
    \centering
    \includegraphics[width=\linewidth]{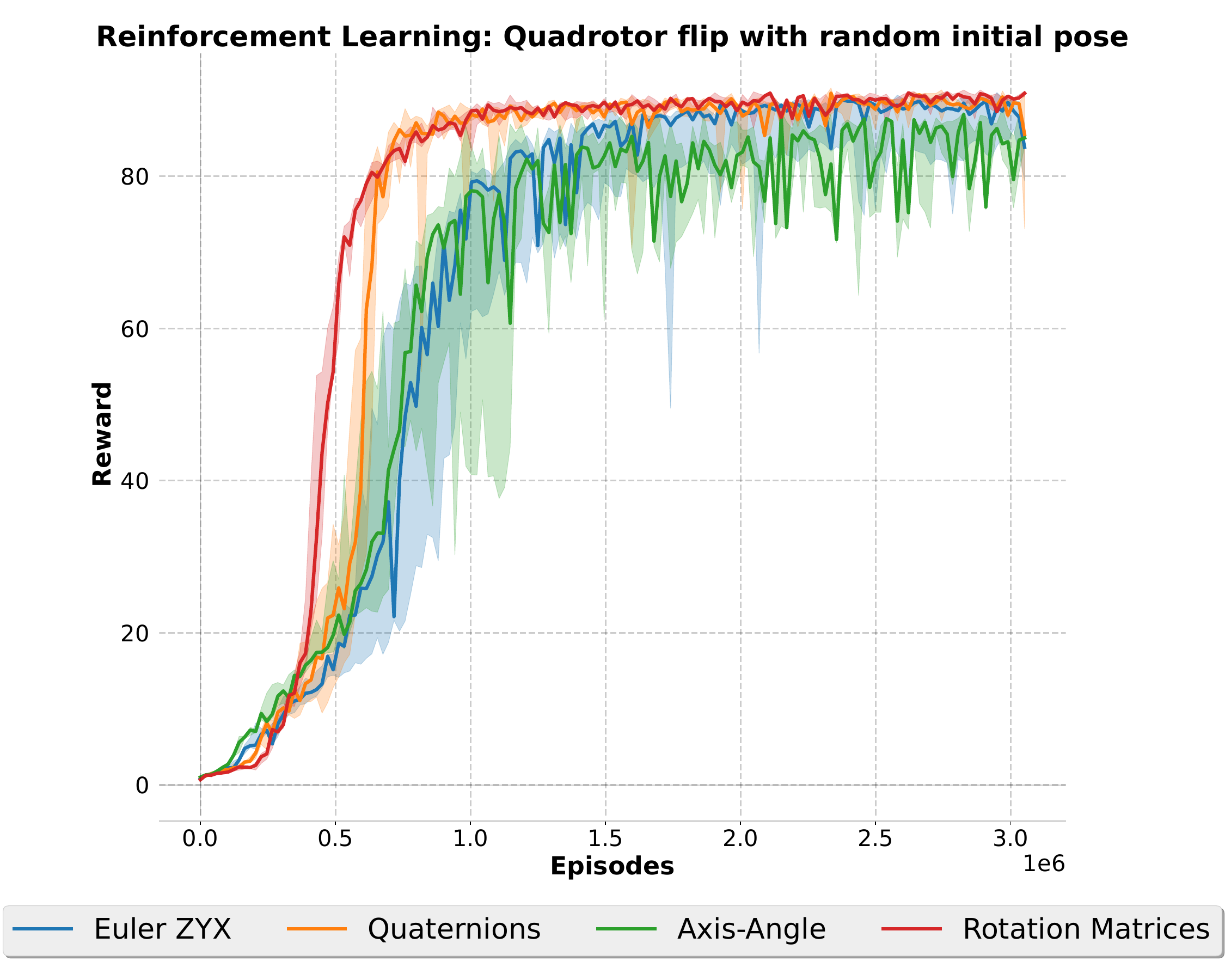}
    \caption{Quadrotor Flip RL Task: During training and evaluation, the starting position of the robot is random. Solid lines are the median over 5 replicates and the shaded regions are the regions between the 25-th and 75-th percentiles.}
    \label{fig:rl_quad_random}
    \vspace{-1em}
\end{figure}
\begin{figure}[!htb]
    \centering
    \includegraphics[width=\linewidth]{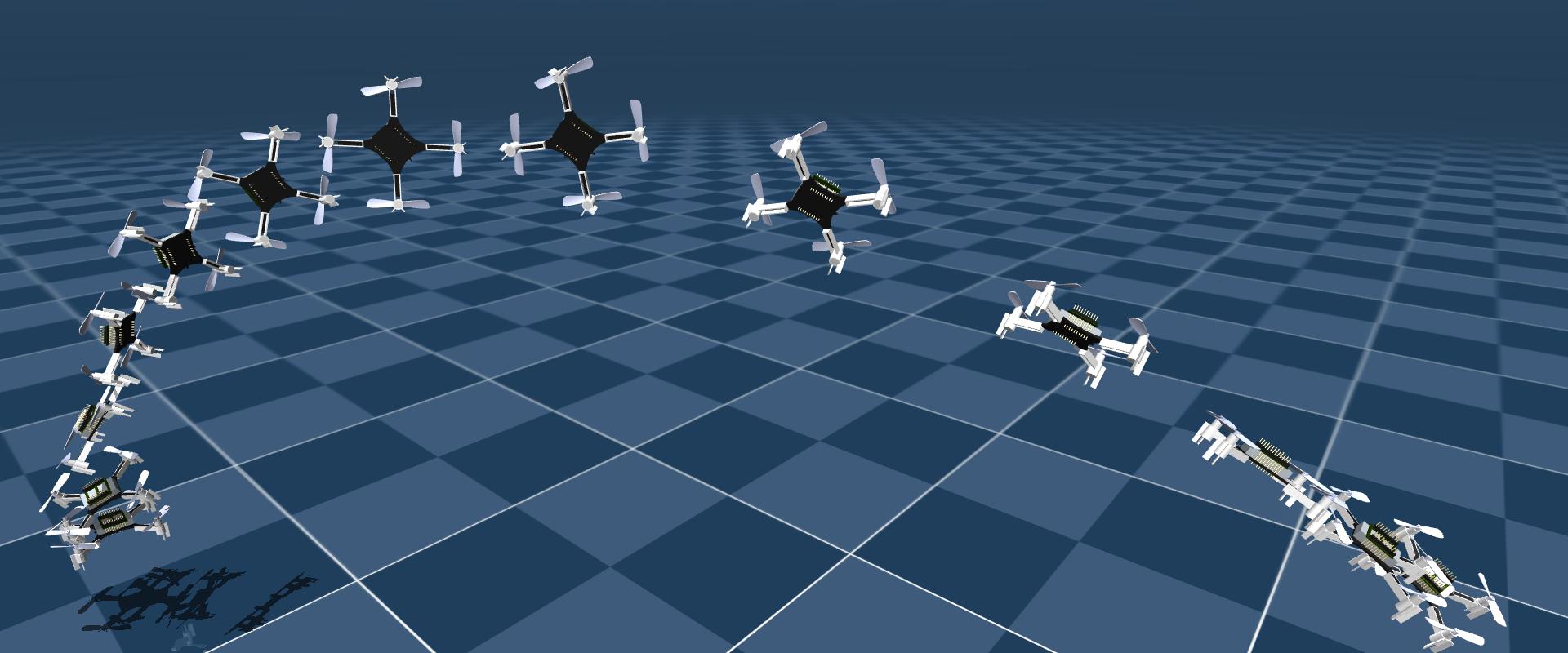}
    \caption{Quadrotor Flip RL Task: The trajectory of policy achieves to perform the flip using the quaternion representation.}
    \label{fig:quad_traj}
    \vspace{-2em}
\end{figure}

We train the policies using a fixed initial state and devise two separate evaluation scenarios. In the first scenario, we evaluate the learned policies with the exact same conditions as the training. Here, the results showcase that all representations are able to learn the task, while the ones that use rotation matrices or Quaternions converge faster and with much less variance (Fig.~\ref{fig:rl_frame_fixed},~\ref{fig:rl_quad_fixed}). In the second scenario we would like to test the generalization capabilities of each representation, and thus during evaluation we start from a random initial orientation (and position for the quadrotor). The results again showcase the superiority of using rotation matrices or Quaternions as the input to the policy (Fig.~\ref{fig:rl_frame_random},~\ref{fig:rl_quad_random}). Qualitatively, all representations are able to learn both tasks, a nice flip trajectory for the quadrotor (Fig.~\ref{fig:quad_traj}).
\begin{messagebox}{msg:rl}
    In Reinforcement Learning settings, we suggest to use Quaternions or \emph{``flattened rotation matrices''} as they provide similar performance.
    ~\\\textbf{Recommended readings:}~\cite{geist2024learning}.
\end{messagebox}
\vspace{-1em}
\subsection{Trajectory Optimization via Differential Dynamic Programming}
Trajectory optimization plays a central role in state-of-the-art methods for generating robot controllers. Here we devise two tasks similar to the RL scenarios. In other words, we have a simple frame that needs to rotate to a desired orientation and a quadrotor that has to perform a flip.

We formulate the problem as follows:
\begin{align}
    \min_{x_{1:N}, u_{1:N-1}} \quad &\ell_f(x_N) + \sum_{k=1}^{N-1} \ell_k(\boldsymbol{x}_k, \boldsymbol{u}_k)\nonumber\\
    \text{subject to} \quad &x_{k+1} = f(\boldsymbol{x}_k, \boldsymbol{u}_k),\nonumber\\
    \quad &g_k(\boldsymbol{x}_k, \boldsymbol{u}_k) \leq 0,\nonumber\\
    \quad &h_k(\boldsymbol{x}_k, \boldsymbol{u}_k) = 0.
\end{align}




Since there is no widely accepted generic optimizer for manifold spaces, we implemented from scratch an iLQR framework for the different representations. We adapted the iLQR algorithm to optimize directly over $\boldsymbol{\Delta x}=\boldsymbol{x}\ominus\bar{\boldsymbol{x}}_k$, and when necessary taking the $SO(3)$ differences. To solve this problem iteratively, we linearize the dynamics around a trajectory $\bar{\boldsymbol{x}}_k,\bar{\boldsymbol{u}}_k$ and approximate the cost function quadratically:
\begin{align}
    & \boldsymbol{\Delta x}_{k+1} = \boldsymbol{A}_k\boldsymbol{\Delta x}_k + \boldsymbol{B}_k \boldsymbol{\Delta u}_k,\nonumber\\
    \Delta J &= \frac{1}{2} \begin{bmatrix} \boldsymbol{\Delta x}_k \\ \boldsymbol{\Delta u}_k \end{bmatrix}^T
    \begin{bmatrix} \boldsymbol{Q}_{xx} & \boldsymbol{Q}_{xu} \\ \boldsymbol{Q}_{ux} & \boldsymbol{Q}_{uu} \end{bmatrix}
    \begin{bmatrix} \boldsymbol{\Delta x}_k \\ \boldsymbol{\Delta u}_k \end{bmatrix}\nonumber\\&\quad\quad+ \boldsymbol{Q}_x^T \boldsymbol{\Delta x}_k + \boldsymbol{Q}_u^T \boldsymbol{\Delta u}_k
\end{align}
Considering $ Q_k(\boldsymbol{x},\boldsymbol{u})=l(\boldsymbol{x},\boldsymbol{u})+V_{k+1}(\boldsymbol{x})$, we have (ignoring second order derivatives, aka iLQR):
\begin{align}
    \boldsymbol{Q}_x &= \nabla_x \ell + \boldsymbol{A}_k^T \boldsymbol{p}_{k+1}\nonumber\\
    \boldsymbol{Q}_u &= \nabla_u \ell + \boldsymbol{B}_k^T \boldsymbol{p}_{k+1}\nonumber\\
    \boldsymbol{Q}_{xx} &= \nabla^2_{xx} \ell + \boldsymbol{A}_k^T \boldsymbol{P}_{k+1} \boldsymbol{A}_k\nonumber\\
    \boldsymbol{Q}_{uu} &= \nabla^2_{uu} \ell + \boldsymbol{B}_k^T \boldsymbol{P}_{k+1} \boldsymbol{B}_k\nonumber\\
    \boldsymbol{Q}_{xu} &= \nabla^2_{xu} \ell + \boldsymbol{A}_k^T \boldsymbol{P}_{k+1} \boldsymbol{B}_k\nonumber\\
    \boldsymbol{Q}_{ux} &= \nabla^2_{ux} \ell + \boldsymbol{B}_k^T \boldsymbol{P}_{k+1} \boldsymbol{A}_k
\end{align}
By Bellman’s principle, the optimal feedback policy is:
\begin{align}
    \boldsymbol{k}_k &= \boldsymbol{Q}_{uu}^{-1} \boldsymbol{Q}_u, \quad \boldsymbol{K}_k = \boldsymbol{Q}_{uu}^{-1} \boldsymbol{Q}_{ux}\nonumber\\
    \boldsymbol{u}_k&= \bar{\boldsymbol{u}}_k - \boldsymbol{k}_k - \boldsymbol{K}_k \boldsymbol{\Delta x}_k
\end{align}

and approximating the value function as a quadratic, $V_k(\boldsymbol{x}+\boldsymbol{\Delta x})=V_k(\boldsymbol{x})+\boldsymbol{p}_k^T\boldsymbol{\Delta x}+\frac{1}{2}\boldsymbol{\Delta x}^T \boldsymbol{P}_k \boldsymbol{\Delta x}$, we can compute quadratic expansion of the cost-to-go:
\begin{align}
\boldsymbol{P}_k=\boldsymbol{Q}_{xx}+\boldsymbol{K}_k^T\boldsymbol{Q}_{uu}\boldsymbol{K}_k-\boldsymbol{Q}_{xu}\boldsymbol{K}_k-\boldsymbol{K}_k^T\boldsymbol{Q}_{ux}\nonumber\\
\boldsymbol{p}_k=\boldsymbol{Q}_{x}-\boldsymbol{K}_k^TQ_{u}+\boldsymbol{K}_k^T\boldsymbol{Q}_{uu}\boldsymbol{k}_k-\boldsymbol{Q}_{ux}\boldsymbol{k}_k
\end{align}
The process iterates until convergence. In our implementation, we perform line-search with restarts and regularization~\cite{jackson2019ilqr,mastalli2022feasibility}, and we have implemented the ``BoxQP'' method for handling control limits~\cite{tassa2014control}.

\textbf{Scenarios:}
We compare the following representations: Euler Angles, Axis Angles, Quaternions, Quaternions with attitude Jacobian and working directly on $SO(3)$. We perform the following experiments:
\begin{enumerate}
    \item A simple frame starts from a random initial orientation and has to reach a random final orientation in $T=4\,s$ with $dt=0.02\,s$. The state space is $\boldsymbol{x} = \begin{bmatrix}\boldsymbol{R}&\boldsymbol{\omega}\end{bmatrix}^T$ and the action space is $\boldsymbol{u}=\dot{\boldsymbol{\omega}}$. \textbf{We perform 200 random scenarios}: that is 200 random initial and final configurations, and random initial $\boldsymbol{u}_k$.
    \item A 3D quadrotor that has performs a flip in $T=2\,s$ with $dt=0.01\,s$. We use the same state space as in the previous sections, and create a cost function that enables the flip. \textbf{We perform 20 scenarios} with different random initialization to the solver (that is, different initial $\boldsymbol{u}_k$).
\end{enumerate}
With the first experiment we will be able to purely validate the effect the orientation representation has on the optimization convergence, while the second experiment will showcase whether this effect transfers to more difficult problems.
\begin{figure}[!htb]
    \centering
    \includegraphics[width=\linewidth]{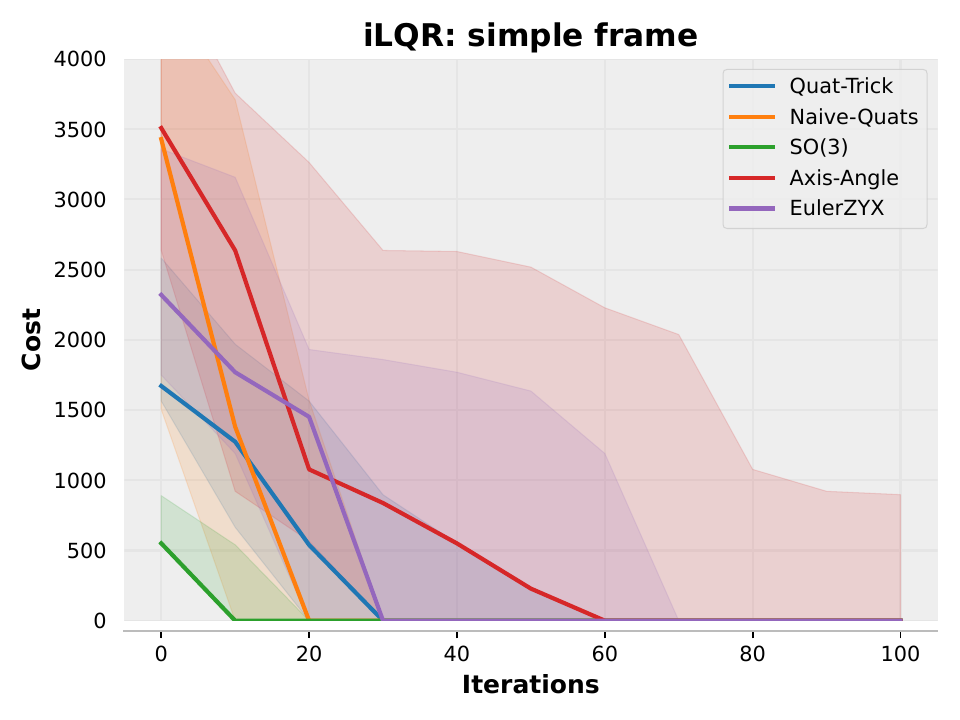}
    \vspace{-2em}
    \caption{iLQR simple frame performance. Solid lines are the median over 200 replicates and the shaded regions are the regions between the 25-th and 75-th percentiles.}
    \label{fig:ilqr_simple}
    \vspace{-1.5em}
\end{figure}
\begin{figure}[!htb]
    \centering
    \includegraphics[width=\linewidth]{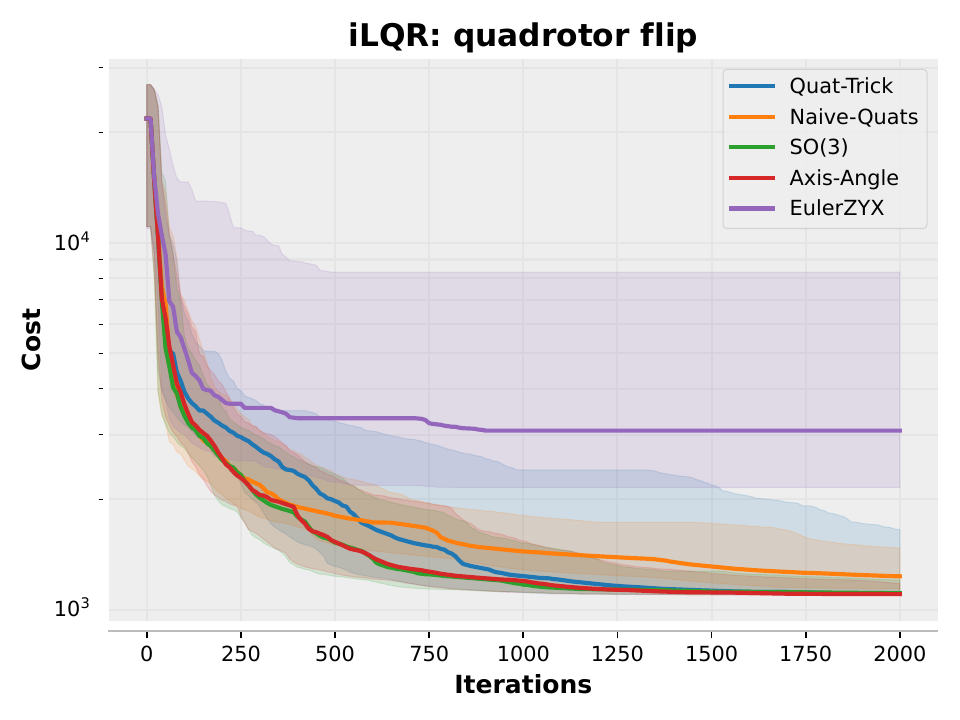}
    \vspace{-2em}
    \caption{iLQR quadrotor flip performance. Solid lines are the median over 20 replicates and the shaded regions are the regions between the 25-th and 75-th percentiles.}
    \label{fig:ilqr_quad}
    \vspace{-2.5em}
\end{figure}

\textbf{Results:} The results showcase that working directly on $SO(3)$ greatly outperforms all other representations. In particular, in the simple frame scenario, using the $SO(3)$ space we always converge in less than 10 iterations no matter the initial and final orientations, while the rest of representations require more iterations and they are dependent on the ``difficulty'' of the specific scenario (Fig.~\ref{fig:ilqr_simple}). Moreover, in the quadrotor flip scenario, when working in $SO(3)$ directly, we always get better total costs and we converge in fewer iterations (Fig.~\ref{fig:ilqr_quad}). Moreover, Axis-Angle also performs quite consistently and finds the same solutions as the $SO(3)$ space most of the time. Quaternions without the Attitude Jacobian can also converge very fast (in wall-time) since each iteration is fast, but can sometimes produce sub-optimal solutions. Euler Angles completely fail in this scenario.
%
%

\begin{messagebox}{msg:ddp}
    In Trajectory Optimization settings, we suggest to use the \emph{$SO(3)$ space and derivatives}. In settings where the underlying numerical optimization algorithm does not that support manifolds, we suggest to use the Axis-Angle representation (aka, the tangent space of $SO(3)$).
    ~\\\textbf{Recommended readings:}~\cite{jallet2023proxddp}.
\end{messagebox}

\vspace{-0.4em}
\section{Conclusion}\label{sec:conclusion}
In this work we provide a systematic overview of orientation fundamentals and a critical survey of the most widely used orientation representations. Our goal is to give roboticists both an intuitive grasp of the underlying geometry and a dependable reference for practical implementation. To expose the relative strengths and limitations of each representation, we evaluate them across different common scenarios --- direct optimization, imitation learning, reinforcement learning, and trajectory optimization.

The results showcase that two representations stand out. For imitation and reinforcement learning scenarios, flattened rotation matrices deliver strong, stable performance; although this might seem counter-intuitive, it is consistent with the literature~\cite{geist2024learning}. In optimization-based scenarios, working directly on the $SO(3)$ manifold --- or if not possible in its tangent‑space (Axis‑Angle) --- yields faster convergence and higher accuracy.


\bibliographystyle{ieeetr}
\bibliography{references}

\end{document}